\documentclass{article}
\usepackage{overpic}
\usepackage{spconf,amsmath,graphicx,import,color}
\graphicspath{ {./images/} }

\title{Deep Perceptual Image Quality Assessment for Compression}
\name{Juan Carlos Mier*\thanks{* equal contribution}, Eddie Huang*, Hossein Talebi, Feng Yang, and Peyman Milanfar}
\address{Google Research}

\begin{document}
%

\maketitle

\begin{abstract}
Lossy Image compression is necessary for efficient storage and transfer of data. Typically the trade-off between bit-rate and quality determines the optimal compression level. This makes the image quality metric an integral part of any imaging system. While the existing full-reference metrics such as PSNR and SSIM may be less sensitive to perceptual quality, the recently introduced learning methods may fail to generalize to unseen data. In this paper we propose the largest image compression quality dataset to date with human perceptual preferences, enabling the use of deep learning, and we develop a full reference perceptual quality assessment metric for lossy image compression that outperforms the existing state-of-the-art methods. We show that the proposed model can effectively learn from thousands of examples available in the new dataset, and consequently it generalizes better to other unseen datasets of human perceptual preference. The CIQA dataset can be found at https://github.com/google-research/google-research/tree/master/CIQA 
\end{abstract}
\begin{keywords}
Full-reference Image Quality Assessment, Image Compression, Compression Quality Metric, Perceptual Quality Dataset
\end{keywords}

\section{Introduction}
\label{sec:intro}

Compressed image data make up a very large portion of data stored in data centers around the world. At the rapidly increasing scale of data storage, compression techniques are more important now than ever. Even a marginal change in efficiency can have large impact on data storage. While image compression techniques are vital for efficiently storing enormous quantities of data, they have remained relatively static over the years \cite{jpeg_alg}. 

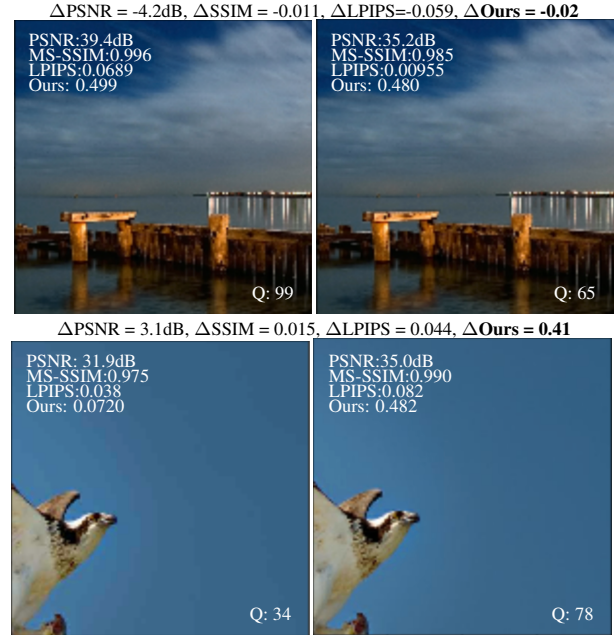
\begin{figure}[t]
    \vspace{-0mm}
    \centering
    \scriptsize
    {$\Delta$PSNR = -4.2dB, $\Delta$SSIM =  -0.011, $\Delta$LPIPS=-0.059, \textbf{$\Delta$Ours = -0.02}}
    
    \begin{overpic}[width=.46\linewidth]{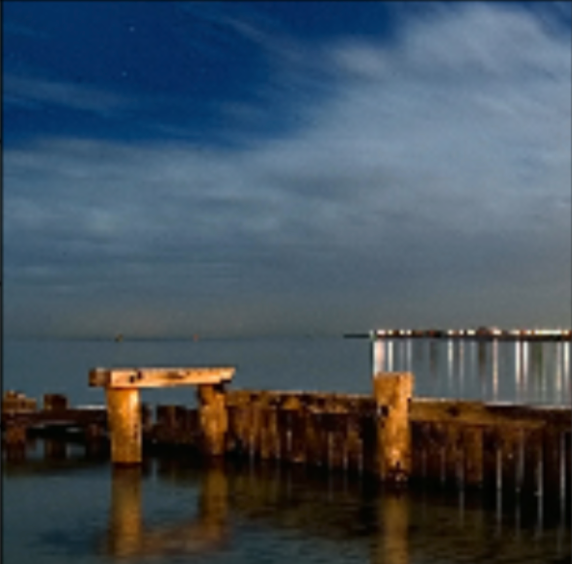}
    \put(5,90){\begin{color}{white}PSNR:39.4dB\end{color}}
    \put(5,85){\begin{color}{white}MS-SSIM:0.996\end{color}}
    \put(5,80){\begin{color}{white}LPIPS:0.0689 \end{color}}
    \put(5,75){\begin{color}{white}Ours: 0.499\end{color}}
    \put(80,5){\begin{color}{white}Q: 99\end{color}}
    \end{overpic}
    \begin{overpic}[width=.46\linewidth]{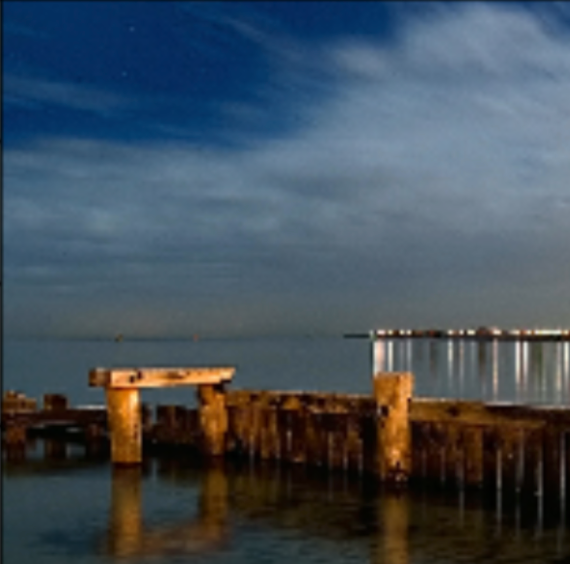}
    \put(5,90){\begin{color}{white}PSNR:35.2dB\end{color}}
    \put(5,85){\begin{color}{white}MS-SSIM:0.985\end{color}}
    \put(5,80){\begin{color}{white}LPIPS:0.00955 \end{color}}
    \put(5,75){\begin{color}{white}Ours: 0.480\end{color}}
    \put(80,5){\begin{color}{white}Q: 65\end{color}}
    \end{overpic}
    
    {$\Delta$PSNR = 3.1dB, $\Delta$SSIM =  0.015, $\Delta$LPIPS = 0.044, \textbf{$\Delta$Ours = 0.41}}
    
    \begin{overpic}[width=.46\linewidth]{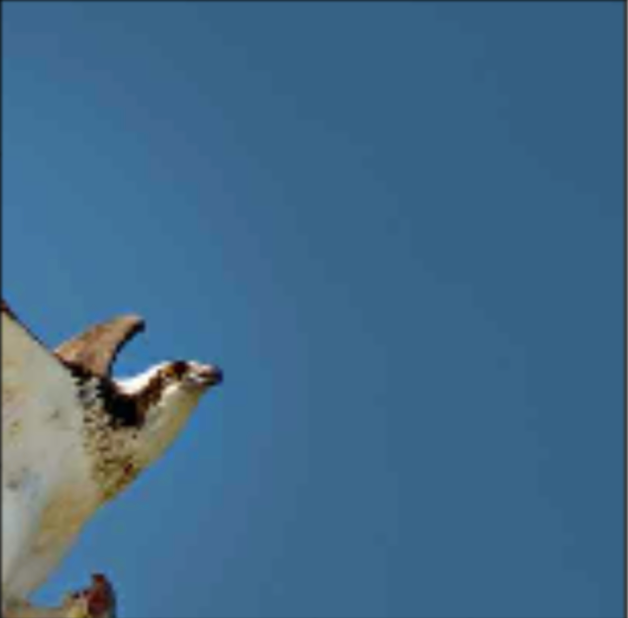}
    \put(5,90){\begin{color}{white}PSNR:  31.9dB\end{color}}
    \put(5,85){\begin{color}{white}MS-SSIM:0.975\end{color}}
    \put(5,80){\begin{color}{white}LPIPS:0.038 \end{color}}
    \put(5,75){\begin{color}{white}Ours: 0.0720\end{color}}
    \put(80,5){\begin{color}{white}Q: 34\end{color}}
    \end{overpic}
    \begin{overpic}[width=.46\linewidth]{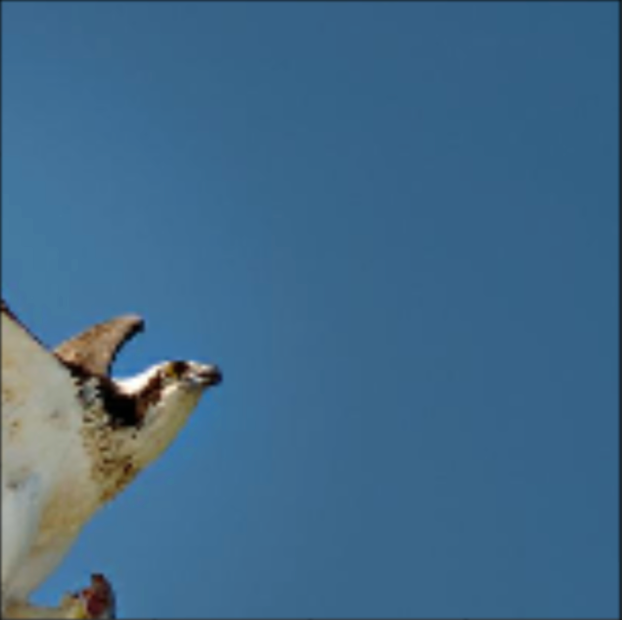}
    \put(5,90){\begin{color}{white}PSNR:35.0dB\end{color}}
    \put(5,85){\begin{color}{white}MS-SSIM:0.990\end{color}}
    \put(5,80){\begin{color}{white}LPIPS:0.082 \end{color}}
    \put(5,75){\begin{color}{white}Ours: 0.482\end{color}}
    \put(80,5){\begin{color}{white}Q: 78\end{color}}
    \end{overpic}
    \vspace{.0em}
    
\caption{Examples of CIQA pairwise comparisons with labeled JPEG Q factor, PSNR, SSIM, LPIPS and our model output. PSNR and SSIM do not adequately correlate with visible distortions in the compressed image. For each metric the $\Delta$ is taken as $Metric(Image_2, Reference) - Metric(Image_1, Reference)$.}
    \label{fig:examples}
\end{figure}

Lossy image compression is a family of techniques for digital images, where bits are discarded when compression takes place resulting in a possible degradation in quality perceived by a human. The degree of compression or bit rate is directly proportional to file size. There is an inherent trade-off between the bit-rate and distortion of a compressed image \cite{qTables}. Currently, some of the the main metrics used for adjusting bit-rate of an encoder is MSE (mean squared error), PSNR (peak-signal-to-noise-ratio), or other similar quality metrics. However, it has been shown that large PSNR values do not necessarily mean a high perceptual quality. For instance, the JPEG encoder can generate artifacts such as color banding or blockiness, that are not well-captured by PSNR \cite{tid2013}. 
In addition, other popular baseline metrics such as SSIM~\cite{wang2004image} and MS-SSIM~\cite{msssim} do not show a consistently high correlation with human perceptual quality preferences. Assessing compression quality based on these metrics can lead to sub-par performances \cite{understanding_ssim}. This leads to low confidence in the results of compression with adjusted bit rates or to the use of a catch-all bit-rate for the encoder to minimize the perceptual quality loss for the vast majority of images \cite{understanding_ssim}.

Learning based methods are the logical next step. These methods do show a significant improvement on correlation with human preferences \cite{zhang2018unreasonable,nima}. The LPIPS metric~\cite{zhang2018unreasonable} has very promising results correlating with human perceptual preferences in measuring similarity of images that have very different and obvious perturbations. However, since most of these models are developed to address generic quality assessment, they may not perform well outside of their main objective and do not generalize consistently to unseen data.

This creates need for a perceptual quality metric that can be used for image compression. Such a metric should not only correlate well with human perceptual preferences but also generalize well to new datasets, robust to unseen data in the wild. To this aim, we first introduce the largest set of human perceptual preference labels for compressed image pairwise comparisons that is better suited for the creation of deep learning networks for compression image quality assessment (IQA). Since 72.4\% of  all  websites  on  the  internet  use  images  with JPEG  standard \cite{jpeg_usage}, we focus on this compression format. Our proposed CIQA is $3\times$ larger than previous datasets enabling semantic information to be more general with more examples and allows the model to focus on compression artifacts during training. Secondly, this paper introduces a deep learning full-reference, perceptual quality metric, trained on CIQA for image compression IQA that correlates strongly with human perception and preference. This model not only outperforms current state-of-the-art (SOTA) metrics on the dataset used for training but also generalizes well to SOTA performance with unseen datasets, namely LIVE-Wild~\cite{live} and FG-IQA~\cite{fgiqa}.

\begin{table*}[]
    \centering
    \begin{tabular}{|c|c|c|c|c|}
        \hline 
        
        Database    &   No.  Ref Images  & No. Distortions & No. Distinct Comparisons & Number of Ratings  \\ \hline \hline
        LIVE-Wild       &   29  &   779     &   780     &   single stimulus (20-29 per image)  \\ 
        FGIQA       &   100 &   1,200   &   3600    &   54,000 (15 per pair) \\ 
        CIQA   &   6,667&   13,868   &   7808    &   249,856 (32 per pair) \\
        \hline
    \end{tabular}
    \caption{Summary of current Compression IQA Datsets}
    \label{tab:my_label}
\end{table*}

\section{Related Research}
\label{sec:relatedwork}

In recent years, there has been a resurgence in interest for image compression techniques particularly in IoT and robotics applications \cite{iot1,iot2}, particularly with the advance in machine learning, which introduces the capability for personalized or smart compression as a way to increase efficiency while minimizing perceptual quality loss. 

\subsection{Compression}
Storing digital images is subject to an important tradeoff between compression file size (bit-rate), perceptual quality (distortion) \cite{qTables}. Different domains emphasize the tension between different pairs of these goals.

The most well studied obvious is the bit-rate distortion tradeoff. MSE, PSNR, SSIM, MS-SSIM \cite{msssim} etc., are the dominant metrics used for measuring distortion resulting from compression. Recent work in this domain advances the search for a metric and method for maintaining a high quality after compression \cite{distortion1,distortion2}. In \cite{distortion4_loss} discusses the importance of a perceptual quality metric that can be used to train a deep compression model.


\subsection{Compressed Image Quality Assessment}
Image quality assessment broadly has been accelerated by the release of large scale datasets that include image labels of human perceptual ratings. Data sets such as AVA \cite{ava}, TID2013 \cite{tid2013}, KonIQ-10k \cite{koniq-10k} and WILD \cite{live} have enabled the use of deep learning for this task. Image compression perceptual quality datasets, however, are still somewhat limited in size limiting the use of deep learning due to likely over-fitting.

Two prominent datasets \cite{live,fgiqa} and associated analysis have been performed in image quality assessment (IQA) for compressed images. \cite{live} created the LIVE-Wild in which 100 reference images were compressed to different degrees using different encoders and the compressed images were rated using single stimulus ranking. \cite{fgiqa} introduces the fine-grained image quality assessment dataset, a much larger-scale image database used for fine grained quality assessment noting the preference of human raters to more subtle changes in compression.

\section{Methods}
\label{sec:methods}

\begin{figure}
    \centering
    \includegraphics[width=7.5cm]{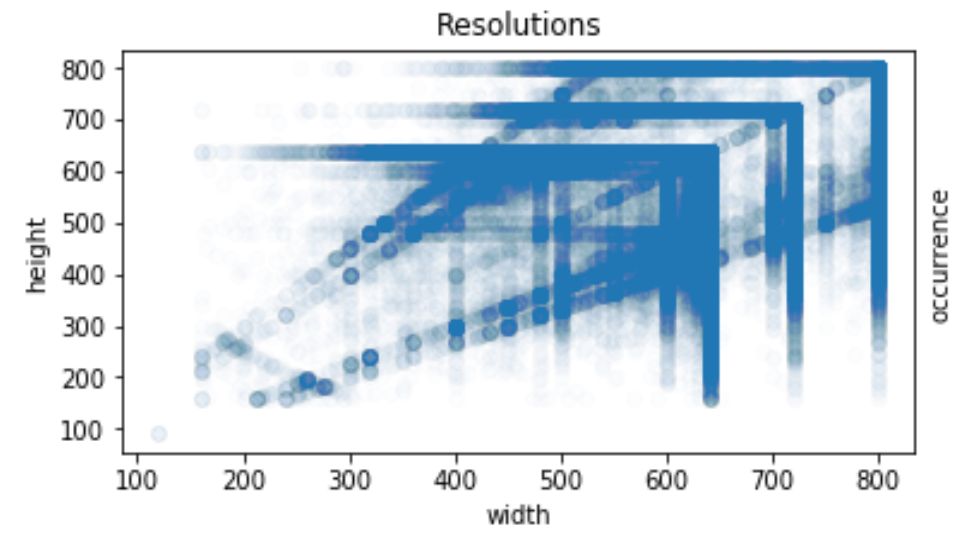}
    \caption{Resolution of images sampled for the CIQA dataset showing diversity in heights and widths}
    \label{fig:image_res}
\end{figure}

The current perceptual quality datasets for image compression, FG-IQA and LIVE-Wild \cite{live,fgiqa}, are well suited for the evaluation of baseline methods, but fall short when training deep learning models due to the limitations in the number of training examples resulting in over-fitting of the network.

The Aesthetic Visual Analysis dataset (AVA) is well suited for deep learning applied to perceptual IQA proven by its successful implementation in the NIMA, NR-IQA model, \cite{nima} and others \cite{usesAva1}. AVA contains $\sim255,000$ images rated based on aesthetic quality by amateur photographers \cite{ava}, and its size and semantic content distribution may be attributed to its success in deep learning applications.  In order to mirror this success to the field of compression perceptual IQA, we introduce the Compression Image Quality Assessment (CIQA) dataset with 6,667 (6400/267 train/test split) unique images sampled from AVA, whose new labels are used to compare quality between images compressed to different JPEG Q values.

Because these images are stored using JPEG compression, only the subset of images with a JPEG Q factor of 99 or more (essentially uncompressed) were sampled and  subsequently compressed at random JPEG quality factors. The original dataset contained semantic labels in addition to perceptual quality labels, and the distribution images from each of the semantic groups was preserved. Semantics labels for each of animal, architecture, cityscape, floral, food/drink, landscape, portrait, and still life each have almost equal occurrence ($\sim6.25\%$) and the category of 'generic' has an occurrence of 50\%.  This was done to ensure the diversity of sampled images and to be a more accurate representation of general perceptual image quality. 

Similarly, Figure \ref{fig:image_res} shows the distribution of resolution from the sampled images which includes ranges from $200\times200$ to $800\times800$, not necessarily always of equal height and width. The distribution of image area $(height \times width)$ followed close to a normal distribution with a mean of $\sim300,000$ pixels and a standard deviation of $\sim100,000$ pixels.

The aforementioned study was a forced choice pairwise comparison study \cite{comparison1, comparison2} and consisted of human raters who were presented with two images derived from the same reference, compressed to different degrees, and were asked to select which image has the better perceptual quality. The rater was asked to select an image at random when unsure. 

This resulted in 7,808 pairwise comparisons were generated and each was rated by 32 individual participants. Pairs were chosen by compressing the reference image to two random quality factors in [10,100]. 13,868 compressed images were generated from the 6,667 reference images, 2 for each image in the train set and 4 for each image in the test set. 

\begin{figure}

\begin{minipage}{1.0\linewidth}
  \centering
  \centerline{\includegraphics[width=9cm]{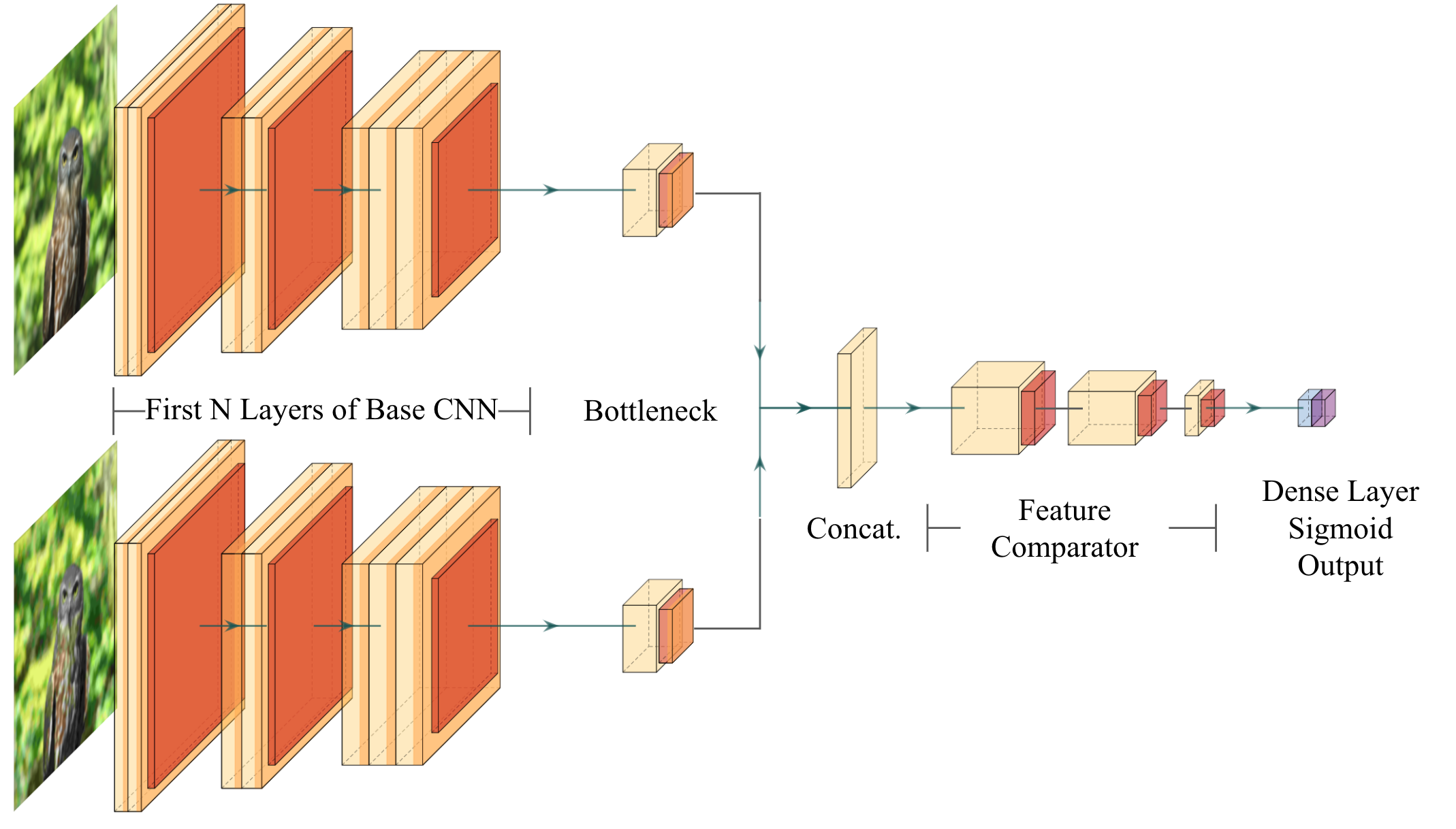}}
  \caption{Architecture used for our proposed deep learning model to predict perceptual IQA preference. The output corresponds to the proportion of raters prefer JPEG2}.
  \label{fig:arch}

  \vspace{-0.805cm}
\end{minipage}
\end{figure}
\section{Perceptual Quality Metric For Compressed Images}
\label{sec:network}


As shown in Figure \ref{fig:arch}, we propose a full-reference CNN model architecture that leverages a partial weight matrix pretrained on ImageNet \cite{imagenet}. The base CNN corresponds to the first $N$ layers from one of, EfficientNet-b0~\cite{efficientnet} (73 layers), EfficientNet-b7 (157 layers), DenseNet~\cite{densenet} (121 layers), ResNet50-v2~\cite{he2016identity} (32 or 44 layers) or VGG16 (7 layers). Both images are fed through a base CNN model to extract features separately up to some some hidden layer l. Both features are then fed through a $1\times1$ convolution filter with 2 features maps to reduce the dimensional of the original features, namely the bottleneck. Next, we concatenate these features, and feed it into the feature comparator, a two layer 1x1 convolutional layer with 256 hidden feature maps and a 1 feature map output. Finally, we apply a global average pooling, an affine transform, and a sigmoid activation, which represents our final prediction. The output is a number between 0 and 1 corresponding to the predicted preference of the second JPEG. A value of 0.5 corresponds to an indistinguishable change in perceptual quality between the two images, and a prediction approaching 0 or 1 corresponds to the respective image being of higher quality.

The neural network based models this architecture described above are trained on our CIQA labels and then evaluated on FG-IQA and LIVE-Wild datasets. Then the current best methods used pervasively including PSNR, MS-SSIM and Q2StepQA are used on the same evaluation set of the three datasets. 

Implementations of PSNR, MS-SSIM and NIQE differ slightly with their implementations in their respective papers as we used the readily available TensorFlow implementation where available. This table includes Q2StepQA which is equal to MS-SSIM multiplied with the negative inverse of NIQE \cite{live}. 

\section{Experimental Results}
\label{sec:typestyle}

The proposed CNN architecture was trained on the new augmented CIQA dataset using multiple base models and subsequently evaluated on FG-IQA and LIVE-Wild to test generalizability in the wild. The results of the CNN model comparison are presented in Table \ref{tab:base_cnn_results}. $N$ layers of the base CNN model were integrated into our architecture. EfficientNet was the best performing CNN by a considerable margin, and the EfficientNet-b0's first 73 layers was chosen due to its performance and lesser complexity when compared to EfficientNet-b7's first 159 layers, also tested.
\begin{table}[]
    \centering
    \begin{tabular}{|c|c|c|c|}
         \hline 
         Models &    LIVE-WILD&   FG-IQA    &   CIQA \\
         \hline \hline
         PSNR       &   0.51    &   -0.29   &   -0.35        \\
         MS-SSIM    &   0.76    &   0.81    &   0.44       \\
         Q2StepQA   &   \textbf{0.79}    &   0.53     &   0.41        \\
         LPIPS      &    0.70       &     0.55      &        0.81       \\
         \textbf{Ours}       &   0.76    &   \textbf{0.89}    &   \textbf{0.91}\\
         \hline 
    \end{tabular}
    \caption{Pearson correlation coefficient of popular and best performing baseline models on Major compression IQA datasets. Ours corresponds to the neural network model trained on CIQA and evaluated on all three datasets.}
    \label{tab:correlations}
\end{table}
\begin{table}[]
    \centering
    \begin{tabular}{|c|c|c|}
    \hline
    Base CNN & Train correlation   &   Test correlation    \\
    \hline \hline
    EfficientNet-b7 & 0.95 & 0.92   \\
    EfficientNet-b0  & 0.90 & 0.88 \\
    DenseNet121 & 0.85 & 0.85   \\
    ResNet50-v2 & 0.87 & 0.82 \\
    VGG16       &0.88 & 0.87 \\
    \hline

    \end{tabular}
    \caption{Pearson correlation coefficient results from various base CNN models}
    \label{tab:base_cnn_results}
\end{table}

The two most important criteria for a compression IQA metric are high correlation with human labels (Accurate) and consistent results across a diverse set of unseen examples (Reliable). For this purpose we use the Pearson Correlation Coefficient. The baseline models show high variability in correlation between datasets and, in some cases, low correlation with labels. Both these qualities indicate a poor real correlation with human perception and a real opportunity for improvement. Table \ref{tab:correlations} shows the correlation of our proposed model using the efficientnet-b0 backbone, trained on CIQA and subsequently evaluated on both CIQA, FG-IQA, and LIVE-Wild compared with the other baseline models tested.

Our proposed model outperforms SOTA for the FG-IQA and our CIQA datasets while remaining competitive when evaluated on the LIVE-Wild dataset achieving similar to MS-SSIM and Q2stepQA. 

The LIVE-Wild dataset used single stimulus scores and the correlation is computed from the difference in the mean opinion scores (DMOS) for the images in the pairwise comparisons giving the labels a different meaning and uses more noticeably compressed images. The model as well as FGIQA and CIQA have labels that represent which is the higher quality image and the degree to which the distinction is clear. The LIVE-Wild labels use a ranking system such that even if the degradation in one image is clear and there is a clear "higher quality image" the magnitude of of the DMOS score depends on more than the two images being compared. The CIQA trained model rates the noticeably distorted images all very low and rates the minimally distorted images in a more linear fashion, which explains the lower correlation with the datasets labels, which is linear across the whole set.


The use of this neural network approach increases the time cost per pixel, but there remains a high potential for optimization. The following inference time is reported for our TensorFlow implementation on an Intel(R) Xeon(R) CPU at 2.30GHz. PSNR: 0.0144 ms/kpixel, LPIPS: 0.215 ms/kpixel, SSIM 0.763 ms/kpixel and our model: 0.675 ms/kpixel.

Our model's input image size of $400\times400$ was produced using random cropping with the possibility for zero padding along dimensions smaller than 400 pixels which is standard in network based image processing. Training was conducted for 15 epochs of 150 steps each with a batch size of 64 and at a learning rate of 0.001. $L_2$ regularization, weight decay, and dropout were not successful in increasing generalizability and were not used.

\section{Discussion}
\label{sec:majhead}

In this work we compare with LPIPS, a SOTA perceptual distance metric, and we evaluate it as a perceptual quality metric, which may not be its intended use. However, the variability seen between datasets is noteworthy. This model is based on human perception and its variance across three datasets supports the need for more training data. 

The proposed model performs at or above SOTA for the three evaluated datasets. The performance of the model trained on the new CIQA data fulfils the two requirements maximizing correlation with perceptual preference and reducing variance across unseen data.

Our proposed model is not necessarily a compression artifact detection algorithm and dataset. The model performs well but not at SOTA when detecting compression artifacts that are difficult for humans to detect. This fact may be a benefit rather than a limitation. Figure \ref{fig:examples} shows in its first example how a large Q value difference can result in an indistinguishable perceptual result ($\sim0.5$) even though the two images differ greatly in file size. This may lead to the potential to use our model as a loss function for the degree of compression or to measure the rate/distortion trade-off.

\section{Conclusion}
The new CIQA dataset is the largest available dataset currently available, opening the possibility for further improvements in deep learning applications to compression IQA. This new augmented database performs significantly better than the current standards when training deep neural networks, leading to less substantial over-fitting and greater generalization. Using this data we were able to train a deep model that outperformed SOTA models across 3 datasets.


\bibliographystyle{IEEEbib}
\bibliography{myBib.bib}

\end{document}